\documentclass[10pt,twocolumn,letterpaper]{article}

\usepackage{cvpr}
\usepackage{times}
\usepackage{epsfig}
\usepackage{graphicx}
\usepackage{amsmath}
\usepackage{amssymb}
\usepackage{graphicx}
\usepackage{subcaption}
\usepackage{caption}
\usepackage{balance}
\usepackage{tablefootnote}

\usepackage[pagebackref=true,breaklinks=true,letterpaper=true,colorlinks,bookmarks=false]{hyperref}

\newcommand{\argmax}{\operatornamewithlimits{argmax}}

\cvprfinalcopy 

\def\cvprPaperID{142} 

\ifcvprfinal\fi
\begin{document}

\title{Beyond Frontal Faces: Improving Person Recognition Using Multiple Cues}

\author{Ning Zhang$^{1, 2}$, \;  Manohar Paluri$^2$, \; Yaniv Taigman$^2$, \; Rob Fergus$^2$, \; Lubomir Bourdev$^2$\\
$^1$UC Berkeley
~~~~~~~~~~~~~~~~~~~
$^2$Facebook AI Research\\
{\tt\small \{nzhang\}@eecs.berkeley.edu} \:\:\:\:
{\tt\small \{mano, yaniv, robfergus, lubomir\}@fb.com}
}

\maketitle

\begin{abstract}
We explore the task of recognizing peoples' identities in  photo albums in an unconstrained setting. 
To facilitate this, we introduce the new  {\em People In Photo Albums (PIPA)} dataset, consisting of over 60000 instances of $\sim$2000 individuals
collected from public Flickr photo albums. With only about half of the person images containing a frontal face, the recognition task is very challenging 
due to the large variations in pose, clothing, camera viewpoint, image resolution and illumination. 
We propose the Pose Invariant PErson Recognition (PIPER) method, which accumulates the cues of poselet-level person recognizers trained by deep convolutional networks to discount for the pose variations, combined with a face recognizer and a global recognizer. 
Experiments on three different settings confirm that in our
unconstrained setup PIPER significantly improves on the performance of
DeepFace, which is one of the best face recognizers as measured on
the LFW dataset. 
\end{abstract}

\section{Introduction}

\begin{figure}[t]
\centering
\includegraphics[scale=0.33] {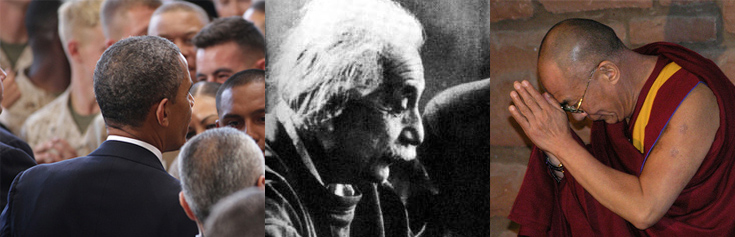}
\caption{We are able to easily recognize people we know in unusual poses, and even if their face is not visible. In this paper we explore the subtle cues necessary for such robust viewpoint-independent recognition. }
\label{fig:teaser}
\end{figure}

Recognizing people we know from unusual poses is easy for us, as illustrated on Figure~\ref{fig:teaser}. In the absence of a clear, high-resolution frontal face, we rely on a variety of subtle cues from other body parts, such as hair style, clothes, glasses, pose and other context. We can easily picture Charlie Chaplin's mustache, hat and cane or Oprah Winfrey's curly volume hair. Yet, examples like these are beyond the capabilities of even the most advanced face recognizers. While a lot of progress has been made recently in recognition from a frontal face, non-frontal views are a lot more common in photo albums than people might suspect. For example, in our dataset which exhibits personal photo album bias, we see that only 52\% of the people have high resolution frontal faces suitable for recognition. Thus the problem of recognizing people from any viewpoint and without the presence of a frontal face or canonical pedestrian pose is important, and yet it has received much less attention than it deserves. We believe this is due to two reasons: first, there is no high quality large-scale dataset for unconstrained recognition, and second, it is not clear how to go beyond a frontal face and leverage these subtle cues. In this paper we address both of these problems. 

\begin{figure}[t]
\centering
\includegraphics[width=0.8\linewidth]{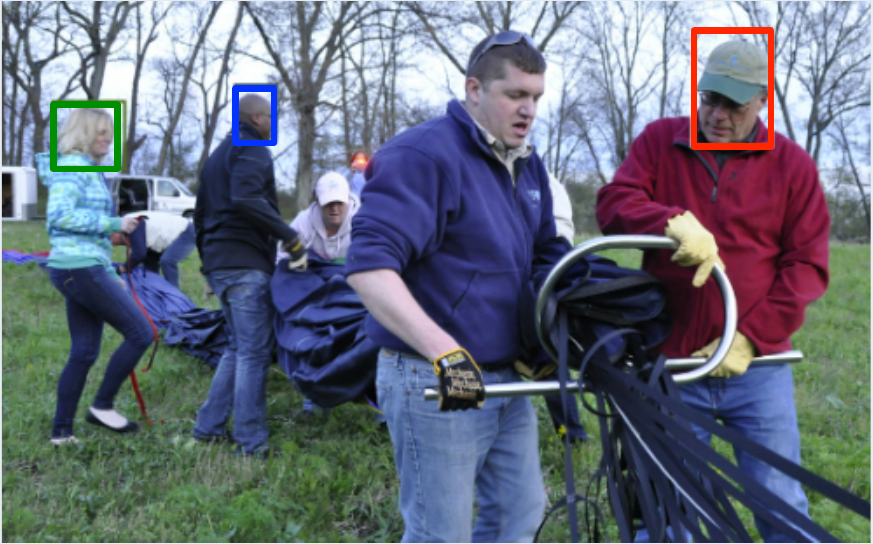}\\
\includegraphics[width=0.5\linewidth]{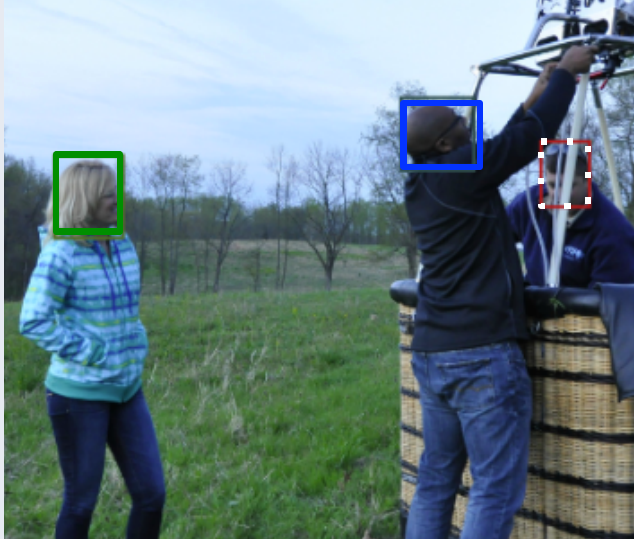}
\includegraphics[width=0.4\linewidth, height=0.43\linewidth]{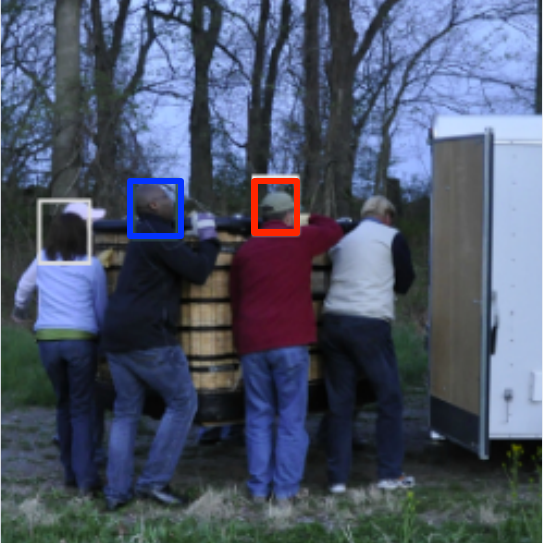}
\caption{\textbf{Example photos from our dataset.}  These are taken from a single album and show the associated identities. Each person is annotated with a ground truth bounding box around the hear, with each color representing one identity. If the head is occluded, the expected position is annotated.} 
\label{fig:dataset_examples}
\end{figure}

\begin{figure*}[t]
\centering
\begin{subfigure}{0.55\textwidth}
\centering
\includegraphics[height=0.4\linewidth]{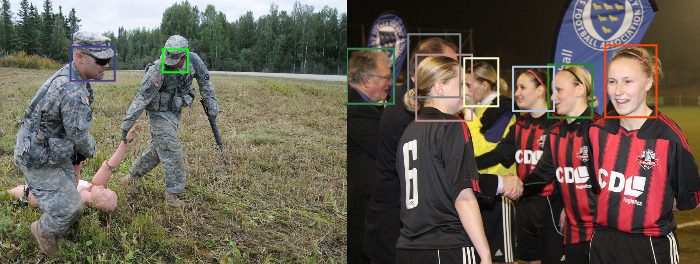}
\caption{While clothing can be discriminative it does not help for military or business activities, for example, where people dress similarly. }
\label{figure:challenges-a}
\end{subfigure}
\begin{subfigure}{0.44\textwidth}
\centering
\includegraphics[height=0.25\textwidth]{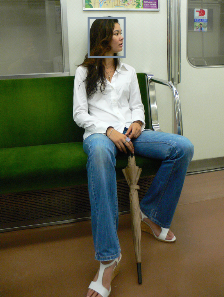}
\includegraphics[height=0.25\textwidth]{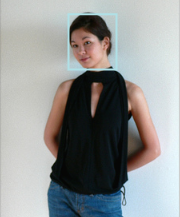} 
\includegraphics[height=0.25\textwidth]{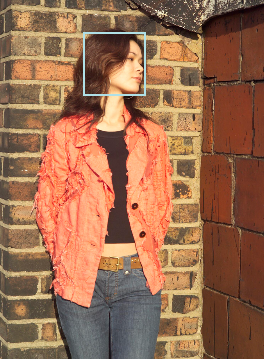} 
\includegraphics[height=0.25\textwidth]{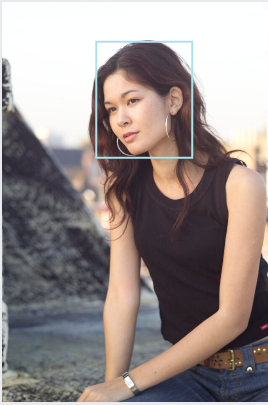} 
\includegraphics[height=0.25\textwidth]{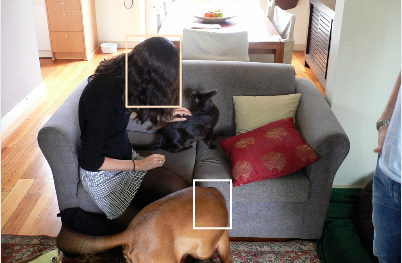} 
\includegraphics[height=0.25\textwidth]{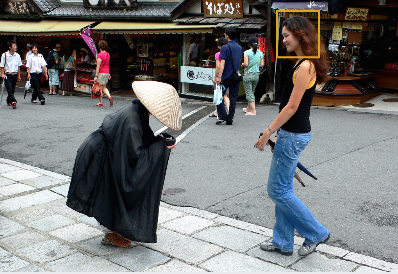} 
\caption{The same individual may appear in multiple albums wearing different clothes.}
\label{figure:challenges-b}
\end{subfigure}
\caption{\textbf{Challenges of our dataset.} Besides significant variations in pose and viewpoint, clothing is not always a reliable cue for person recognition in our dataset.}
\label{fig:challenges}
\end{figure*}
We introduce the {\em People In Photo Albums (PIPA)} dataset, a large-scale recognition dataset collected from Flickr photos with creative commons licenses. It consists of 37,107 photos containing 63,188 instances of 2,356 identities and examples are shown in Figure \ref{fig:dataset_examples}. We tried carefully to preserve the bias of people in real photo albums by instructing annotators to mark every instance of the same identity regardless of pose and resolution. Our dataset is challenging due to occlusion with other people, viewpoint, pose and variations in clothes. While clothes are a good cue, they are not always reliable, especially when the same person appears in multiple albums, or for albums where many people wear similar clothes (sports, military events), as shown in Figure \ref{fig:challenges}.  
As an indication of the difficulty of our dataset, the DeepFace system~\cite{deepface}, which is one of the state-of-the-art recognizers on LFW~\cite{LFWTech}, was able to register only 52\% of the instances in our test set and, because of that, its overall accuracy on our test set is 46.66\%. We plan to make the dataset publicly available.

\begin{figure}[t]
\centering
\includegraphics[width=\linewidth]{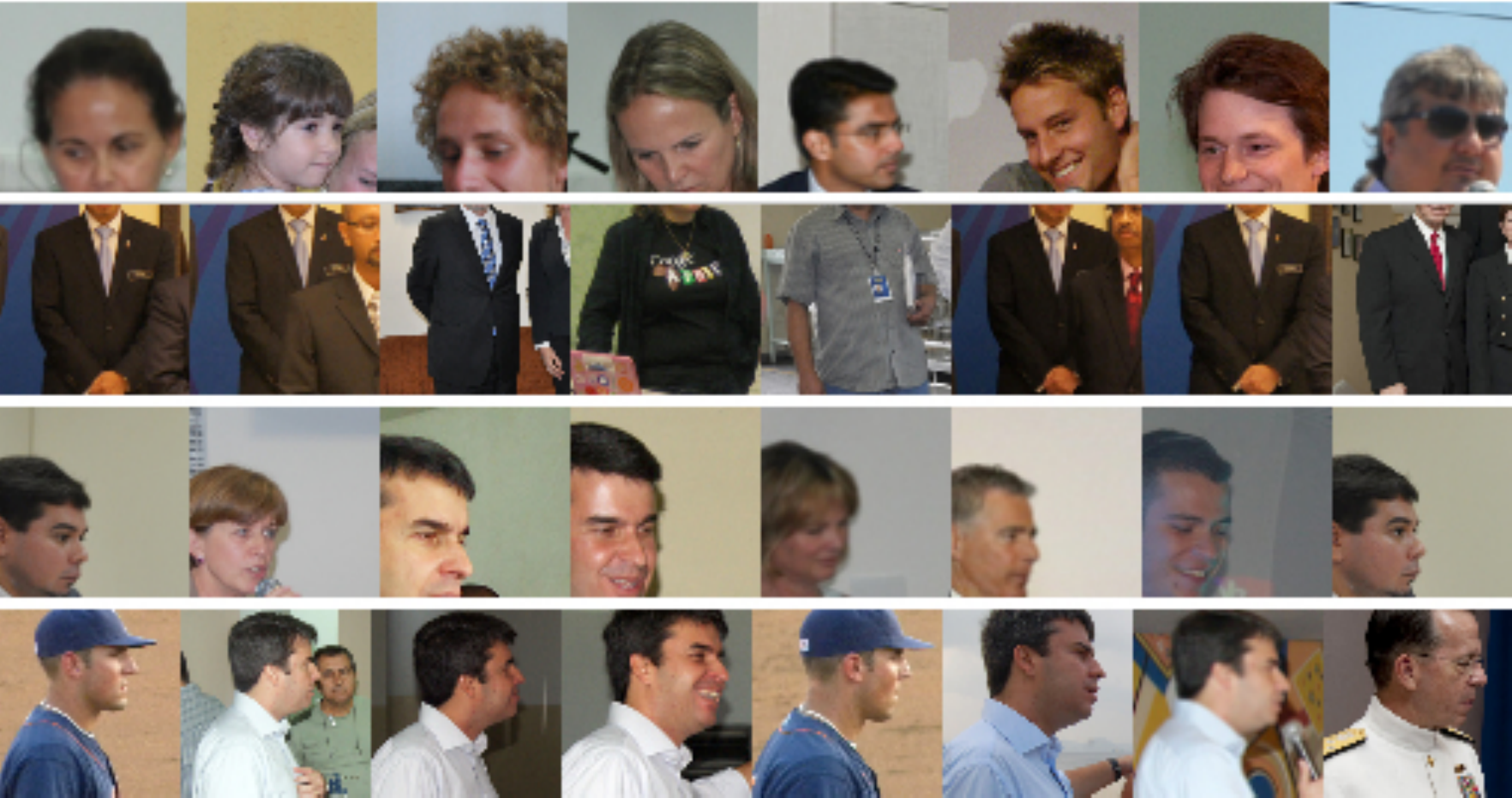}
\caption{Example of poselet activations. These are the top 4 poselet detections ranked by part weight $w_i$ mentioned in Sec \ref{sec:train_w}.}
\label{fig:poselets}
\end{figure}

We propose a Pose Invariant PErson Recognition (PIPER) method, which  uses part-level person recognizers to account for pose variations. 
We use poselets~\cite{Bourdev09}  as our part models and train identity classifiers for each poselet. Poselets are classifiers that detect common pose patterns. A frontal face detector is a special case of a poselet. Other examples are a hand next to a hip or head-and-shoulders in a back-facing view, or legs of a person walking sideways. A small and complementary subset of such salient patterns is automatically selected as described in~\cite{Bourdev09}.
Examples of poselets are shown in Figure \ref{fig:poselets}.
While each poselet is not as powerful as a custom designed face recognizer, it leverages weak signals from specific pose pattern that is hard to capture otherwise. By combining their predictions we accumulate the subtle discriminative information coming from each part into a robust pose-independent person recognition system.

We are inspired by the work of Zhang et al.~\cite{Panda}, which uses deep convolutional networks trained on poselet detected patches for attribute classification. However our problem is significantly harder than attribute classification since we have many more classes with significantly fewer training examples per class. 
We found that combining parts by concatenating their feature in the manner of ~\cite{Panda} is not effective for our task. It results in feature vectors that are very large and overfit easily when the number of classes is large and training examples are few. 
Instead, we 
found training each part to do identity recognition and combining their predictions achieves better performance. Unlike \cite{iccv11}, we propose a new way to handle the sparsity from poselet detections which boosts the performance by a large margin. 

We demonstrate the effectiveness of PIPER by using three different experimental settings on our dataset.
Our method can achieve 83.05\% accuracy over 581 identities on the test set. Moreover when a frontal face is available, it improves the accuracy over DeepFace from 89.3\% to 93.4\%, which is $\sim$40\% decrease in relative error. 

\section{Related Work}

\paragraph{Face recognition}
There has been dramatic progress made in face recognition in the past few decades from EigenFace \cite{eigenface} to the state-of-art face recognition system \cite{deepface} by using deep convolutional nets.
Most of the existing face recognition systems require constrained setting of frontal faces and explicit 3D face alignments or facial keypoint localizations. 
Some other works \cite{robust-facerec, CuiLXSC13} have addressed robust face recognition systems to deal with varying illumination, occlusion and disguise. Due to our unconstrained setting, most of conventional face recognition systems have limited performance on our dataset.

\paragraph{Person identification in photo albums}
Tagging of personal photo albums is an active research topic. To address the limitation of face recognition systems, various approaches have been proposed to incorporate contexts. For examples, the authors in \cite{AnguelovLGS07, naaman}  proposed methods to incorporate contextual cues including clothing appearance and meta data from photos for person identification in photo collections. Sivic et al. \cite{Josef-findpeople} proposed a simple pictorial structure model to retrieve all the occurrences of the same individual in a sequence of photographs. Lin et al. \cite{Lin:2010} presented a generative probabilistic approach to model cross-domain relationships to jointly tag people, events and locations.  
In \cite{GargRSS_CVPR_2011}, the authors try to find all images of each person in the scene on a set of photos from a crowed public event by integrating multiple cues including timestamps, camera pose and co-occurrence of people.

There is also some related work to discover the social connection between people in the photo collections. Wang et al. \cite{Wang:2010} proposed a model to represent the relationship between the social relationships, position and pose of people and their identities. In \cite{Gallagher_understandingimages}, the authors investigated the different factors that are related to the positions of people in a group image. 

 Another interesting direction is to name characters in TV series. In \cite{Everingham:2009, Sivic09, Everingham06a}, the authors proposed approach to automatically label the characters by using aligned subtitle and script text. Tapaswi et al. \cite{tapaswi_knock} proposed a Markov Random Field (MRF) method to combine face recognition and clothing features and they tried to name all the appearance of characters in TV series including non frontal face appearance.  Later they presented another semi-supervised learning  method  \cite{bauml_semi} for the same task. 

\paragraph{Person re-identification in videos}
The task of person re-identification is to match pedestrian images from different cameras and it has important applications in video  
and there are some related work for this topic. Existing work is mainly focused on metric learning \cite{rankSVM, Guillaumin2009iccv2, LiZW12}, mid-level feature learning \cite{viper1, rui1, rui2, FarenzenaBPMC10, layne, Oreifej10humanidentity}.  Li et al. \cite{wei} propose a deep network using pairs of people to encode photometric transformation. Yi et al. \cite{deep-metric-learning} used a siamese deep network to learn the similarity metric between pairs of images. 


\paragraph{Deep convolutional networks}
In the past few years, deep convolutional networks originally pioneered by LeCun et al. \cite{LeCun} have been a tremendous success by achieving the state-of-art performance in image classification \cite{Krizhevsky12}, object detection \cite{rcnn}, face recognition \cite{deepface} and other computer vision tasks.  The strength of the deep nets is its ability to learn discriminative features from raw image input unlike hand-engineered features.  DeCAF \cite{decaf} showed the deep features extracted from the network pretrained on large datasets can be generalized to other recognition problems. 

\section{People In Photo Albums Dataset}
To our knowledge, there is no existing large scale dataset for the task of person recognition. 
The existing datasets for person re-identification, such as VIPeR \cite{viper} and CUHK01 \cite{LiZW12}, come mostly from videos and they are low resolution images taken from different cameras from different viewpoints. The Gallagher Collection Person Dataset~\cite{gallagher_cvpr_08_clothing} is the closest to what we need, however the released subset has only 931 instances of 32 identities which is approximately 1.5\% of the size of our dataset. Furthermore, ~\cite{gallagher_cvpr_08_clothing} have only labeled the frontal faces. The Buffy dataset used in \cite{Sivic09, Everingham:2009} is a video dataset and it only has less than 20 different characters.

Our problem setting is to identify person in the ``wild'' without any assumption of frontal face appearance or straight pedestrian pose. We don't assume that the person is detected by  a detector; our instruction to annotators is to mark the head (even if occluded) for any people they can co-identify, regardless of their pose, and the image resolution.

\subsection{General statistics}
We collected our dataset, People In Photo Albums (PIPA) Dataset, from public photo albums uploaded to Flickr \footnote{\url{https://www.flickr.com/}} and we plan to make it publicly available. All of our photos have Creative Commons Attributions License. Those albums were uploaded from $111$ users. Photos of the same person have been labeled with the same identity but no other identifying information is preserved. Table \ref{table:datasetstats} shows statistics of our dataset. 

\begin{table}
  \centering
  \small
  \begin{tabular}{|c|c|c|c|c|c|}
\hline
Split & All & Train & Val & Test & Leftover\\
\hline
Photos & 37,107 &17,000 & 5,684& 7,868 &6,555\\
Albums & 1,438 & 579& 342&357 &160\\
Instances & 63,188 & 29,223& 9,642& 12,886 & 11,437\\
Identities &2,356 & 1,409& 366& 581& -\\
Avg/identity & 26.82& 20.74&26.34 & 22.18& -\\
Min/identity & 5 &10 &5 & 10&-\\
Max/identity &2928 &99 & 99& 99  &-\\
\hline
\end{tabular}
\caption{Statistics of our dataset.}
\label{table:datasetstats}
\end{table}

%

\subsection{Collection Method}

\begin{figure*}[t]
\centering
\begin{subfigure}{0.5\textwidth}
\includegraphics[width=\linewidth]{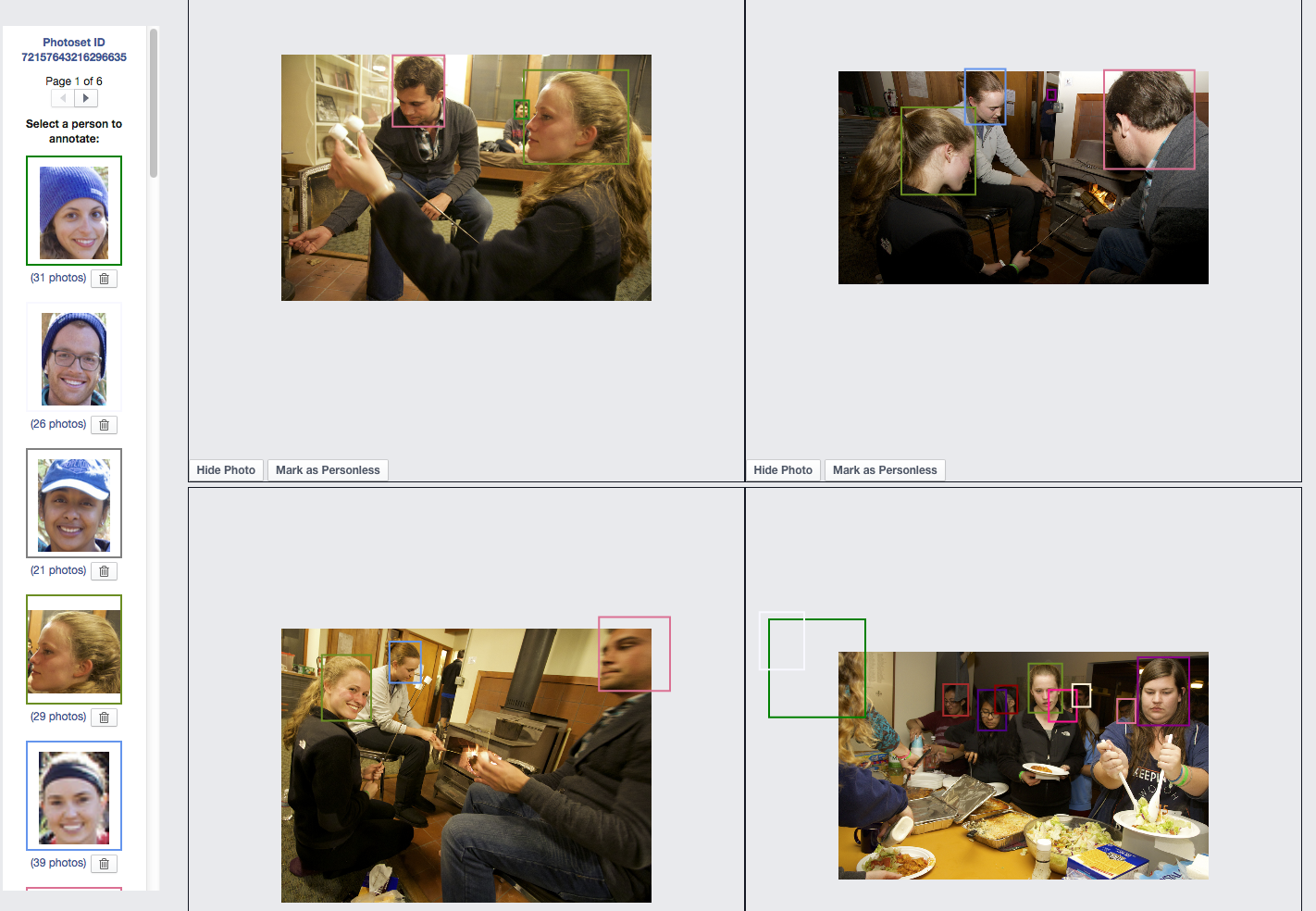}
\caption{Interface for annotating identities in one album.}\label{fig:LabelInterface}
\label{figure:interface-a}
\end{subfigure}
\begin{subfigure}{0.49\textwidth}
\centering
\includegraphics[width=\linewidth]{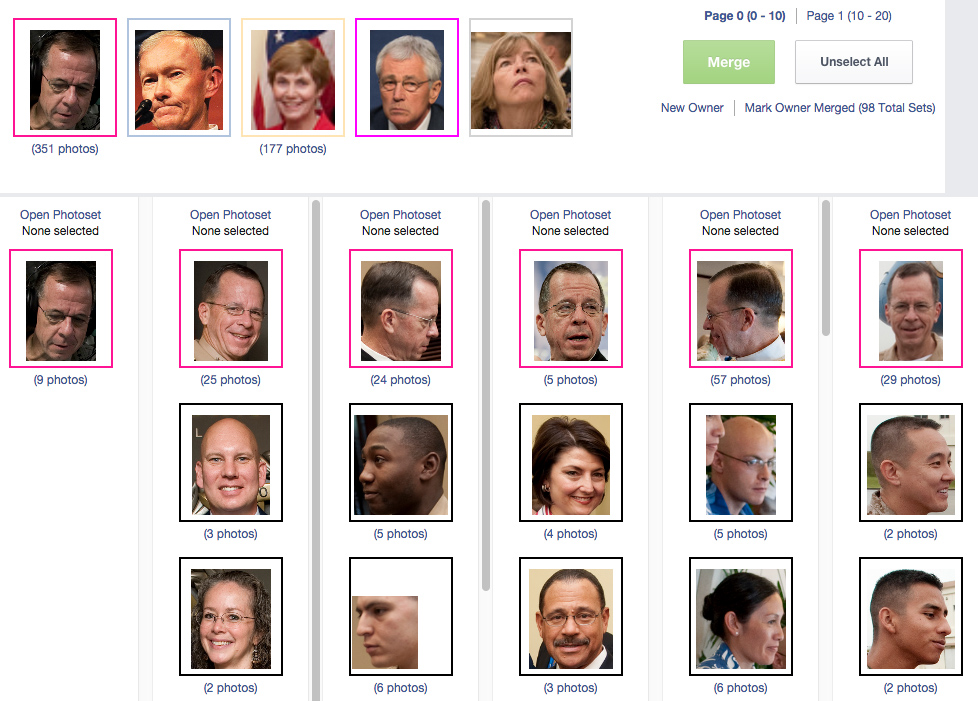}
\caption{Interface for merging identities across albums. }
\label{fig:mergeInterface}
\end{subfigure}
\caption{ \textbf{Interfaces for our annotation system.} The interface for annotating identities in one album in shown in (a) where the annotator can scroll over the faces on the left. If clicking on one individual, it will show all the instances of that person. The original images and head annotations are shown on the right where the annotators draw all the heads of different individuals. If the person's head is occluded, the annotator will draw a bounding box around the expected position of the head.  We show the interface for merging in (b). The top row shows a set of merged individuals. Each column in the bottom section corresponds to an album from the same uploader. Each face is an example face of a different individual. Merging is done by selecting a set of faces across albums, optionally selecting an individual from the top row to merge into, and clicking the merge button.}
\label{figure:interface}
\end{figure*}

Our data collection consists of the following steps:

\begin{enumerate}
\item \textbf{Album Filtering}. After downloading thousands of albums from Flickr, we first show the annotators a set of photos from the album and ask them to filter out albums which are not of people albums, such as landscape, flowers, or photos where person co-occurrence is very low.

\item \textbf{Person Tagging}. Then given each album,  we ask the annotators to select all the individuals that appear at least twice in that album and draw a bounding box around their heads with different color indicating different identity. If the head is partially/fully occluded, we mark the region of where the head should be. The head bounds may also be partially/fully outside the image. Not every person is tagged. In a crowd scene we ask the annotators to tag no more than 10 people. The interface of person tagging is shown in Fig \ref{figure:interface-a}.
\item \textbf{Cross-Album Merging}. Often the same identities appear in multiple albums, in which case their identities should be merged. While it is not feasible to do so across all albums, we consider the set of albums uploaded by the same uploader and we try to find the same identities appearing in multiple albums and merge them. Showing all identities from all albums is a challenging UI task for uploaders that have dozens of large albums. We show our annotation interface in Figure \ref{fig:mergeInterface}. 
\item \textbf{Instance Frequency Normalization}.  After merging, we count the number of instances for each individual and discard individuals that have less than 10 instances. In addition, a few individuals have a very large number of instances which could bias our dataset. To prevent such bias, we restrict the maximum number of instances per individual to be 99. We randomly sample 99 instances and move the remaining ones into a ``leftover" set. Our leftover set consists of 11,437 instances of 54 individuals.
\item \textbf{Dataset Split}. We split the annotations randomly into three sets -- training, validation and test. To ensure complete separation between the sets, all the photos of the same uploader fall in the same set. That ensures that the set of photos, identities and instances across the three sets is disjoint. We do a random permutation of the uploaders and we pick the first $K$ of them so that the number of person instances reaches about 50\% and we assign those to the training set. We assign the next 25\% to validation and the remaining 25\% to test. While we target 50-25-25 split we cannot assure that the instances will be partitioned precisely due to the constraints we impose. See Table~\ref{table:datasetstats} for more details about the splits.
\end{enumerate}

\section{Pose Invariant Person Recognition (PIPER)}

\noindent
We introduce a novel view invariant approach  to combine information of different classifiers for the task of person recognition. It consists of three components:
\begin{itemize}
\item The global classifier, a CNN trained on the full body of the person.
\item A set of 107 poselet classifiers, each is a CNN trained on the specific poselet pattern using~\cite{Bourdev09}. \footnote{The original set of poselets is 150 but some of them did not train well.}
\item An SVM trained on the 256 dimensional features from DeepFace~\cite{deepface}.
\end{itemize}

\noindent In total we have 109 part classifiers. The identity prediction of PIPER is a linear combination of the predicted probabilities of all classifiers:

\begin{equation}
s(X, y) = \sum_i w_i P_i(y|X)
\label{eq:joint}                                                                                                                                                                                                                                                                                                        
\end{equation}

\noindent Here $P_i(y|X)$ is the normalized probability of identity label $y$ given by part $i$ for feature $X$ and $w_i$ is the associated weight of the part. 
The final identity prediction is $y^*(X) = \argmax_ys(X,y)$. 

\noindent Here is an overview of the training steps. The next sections provide a more detailed description.
\begin{enumerate}
\item We run poselets \cite{Bourdev09} over our dataset and match the person predictions coming from poselets to our ground truths (see Section~\ref{sec:detectposelets}).
\item Using the poselet patches of step 1, we train a CNN for each poselet to recognize the identities on our training set. In addition, we train a CNN for the global classifier using the patches corresponding to the full body images. In all cases we use the Convolutional Neural Net architecture by Krizhevsky \etal.~\cite{Krizhevsky12}. We fine-tune the network pre-trained on ImageNet on the task of identity recognition. While recent architectures have improved the state-of-the art \cite{vgg, googlenet} and might further improve our performance, we decided to use the Krizhevsky architecture because its performance is well studied on a number of visual tasks \cite{decaf}. We then discard the final FC8 layer and treat the activations from the FC7 layer as a generic feature on which we train SVMs in the next steps.
\item We partition the validation set into two halves. We train an SVM for each part using the FC7 layer feature from Step 2 on the first half of validation and use it to compute the identity predictions $P_i(y|X)$ on the second half, and vice versa (see Section~\ref{sec:train_p}).
\item We use the identity predictions of all parts on the validation set to estimate the mixing components $w_i$ (see Section~\ref{sec:train_w}).
\item We split the test set in half and train SVMs on top of the FC7 features on the first half of the test set and use them to compute the identity predictions $P_i(y|X)$ on the second half, and vice versa.
\item We use the identity predictions on the test set for each part $P_i(y|X)$ as well as the mixing components $w_i$ to compute the combined identity prediction $S(X,y)$ using equation~\ref{eq:joint}.
\end{enumerate}

In the next sections we will describe the training steps, and way we compute $P_i(y|X)$ and $w_i$.

\subsection{Computing Part Activations}
\label{sec:detectposelets}

\noindent Our groundtruth annotations consist of bounding boxes of heads. 
From the head boxes, we estimate the bounding box locations by setting approximate offset and scaling factor. 
We run poselets~\cite{Bourdev09} on the images, which returns bounding boxes of detected people in the images, each of which comes with a score and locations of associated poselet activations. 
Examples of poselet detections are shown in Figure \ref{fig:poselets}. We use a bipartite graph matching algorithm to match the ground truth bounds to the ones predicted by the poselets. This algorithm performs globally optimal matching by preferring detections with higher score and higher overlap to truths. 
The output of the algorithm is a set of poselet activations associated with each ground truth person instance. 
We extract the image patches at each poselet and use them to train part-based classifiers. 

\subsection{Training the Part Classifiers $P_i(y|X)$ }
\label{sec:train_p}

\subsubsection{Global classifier $P_0(y|X)$}

Using the FC7 layer of the CNN trained for the full body area of each instance, we train a multi-class SVM to predict each identity $y$. We refer to its prediction as $P_0(y|X)$.

\subsubsection{Part-level SVM classifier $\hat{P_i}(y|X)$}
Given the FC7 layer features $X$ extracted from detected part $i$ patch and identity labels $y$,  we train a multi-class SVM on $X$ to predict $y$ and 
we denote the softmax of the output score as $\hat{P_i}(y|X)$. 

Notice that $\hat{P_i}$ is sparse in two ways:
\begin{itemize}
\item Each poselet activates only on instances that exhibit the specific pose of that poselet. Some poselets may activate on 50\% while others on as few as 5\% of the data.
\item Not all identities have examples for all poselets and thus each poselet level SVM classifier for part $i$ is only trained on a subset $F_i$ of all identities. Thus $\hat{P_i}(y|X)$ is inflated when $y\in F_i$ and is zero otherwise. 
\end{itemize}

\noindent The sparsity pattern is correlated with the pose of the person and has almost no correlation to the identity that we are trying to estimate. Thus properly accounting for the sparsity is important in order to get high accuracy identity recognition.


\subsubsection{Sparsity filling}
We address both of these sparsity issues by using the probability distribution of our global model $P_0$ which is defined for all identities and activates on all instances:
\begin{align}
\label{equ:sparsity}
&P_i(y | X) = 
\begin{cases}
P_0(y|X), \text{if part $i$ doesn't activate, or}  \\ 
P(y\in F_i) \hat{P_i}(y | X) + P(y\notin F_i) P_0(y | X)
\end{cases} \\
&P(y\in F_i) = \sum_{y'\in F_i} P_0(y' | X)
\end{align}

\begin{figure}[t]
\centering
\includegraphics[width=\linewidth]{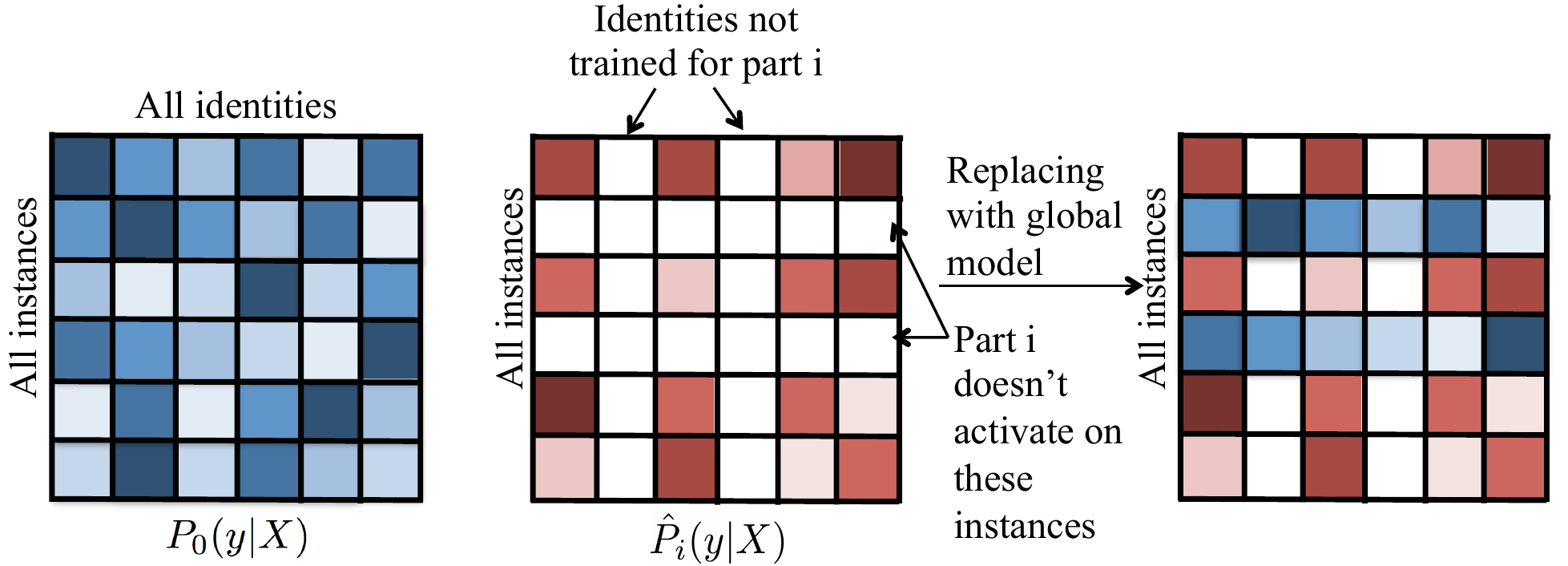}
\caption{Example of sparsity filling. On the left we show the predictions of the global model for every identity and every instance. The poselet classifier in the middle does not activate for two instances (the white rows) and is not trained to recognize two identities (the white columns). On the right we show how in the normalized probability we fill-in missing rows from the global model acc as in the top of Equation \ref{equ:sparsity}. In addition, (not shown in the figure) we account for the missing columns by linearly interpolating each row with the global model based on the likelihood that the identity is not coming from one of the missing columns.}  
\label{fig:sparsity}
\end{figure}

\noindent In Figure \ref{fig:sparsity} we give a visual intuition behind this formula.

\subsection{Computing the part weights $w_i$}
\label{sec:train_w}

We use the validation set to compute $w$. We split the validation set into two equal subsets. We train the part-based SVMs on one subset and use them to compute $P_i(y|X)$ on all instances of the second subset, and vice versa. Let $P_i^j(y|X)$ denote the probability that the classifier for part $i$ assigns to instance $j$ being of class $y$ given feature $X$. We formulate a binary classification problem which has one training example per pair of instance $j$ and label $y$. If we have $K$ parts its feature vector is $K+1$ dimensional: $[P_0^j(y|X); P_1^j(y|X); ~... ~ P_k^j(y|X)]$. Its label is 1 if instance $j$'s label is $y$ and -1 otherwise. We solve this by training a linear SVM. The weights $w$ are the weights of the trained SVM. 

We use the first split of validation to do a grid search for the $C$ parameter of the SVM and test on the second half. Once we find the optimal $C$ we retrain on the entire validation set to obtain the final vector $w$.

\section{Experiments}
We report results of our proposed method on our PIPA dataset and compare it with baselines. 
Specifically, we conduct experiments in three different settings: 1) Person recognition,  2) One-shot person identification, and 3) Unsupervised identity retrieval.


In all experiments we use the training split of our dataset to train the deep networks for our global model and each poselet and we use the validation set to compute the mixing weights $w$ and tune the hyper-parameters. All of our results are evaluated on the test split. 
\subsection{Person Recognition}
\begin{figure*}[t]
\centering
\begin{tabular}{cc}
\centering
\includegraphics[height=0.21\linewidth]{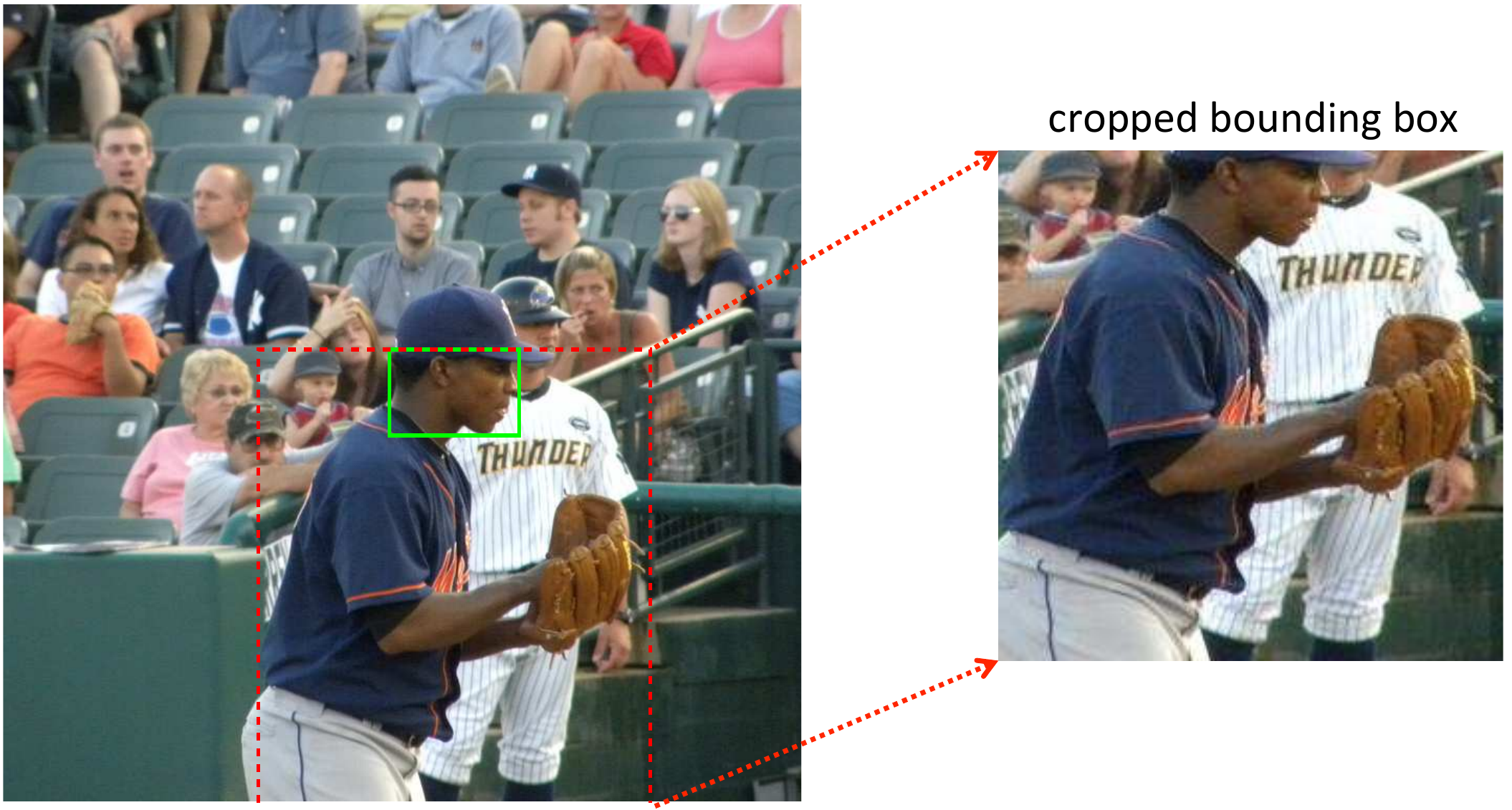}
&
\includegraphics[height=0.21\linewidth]{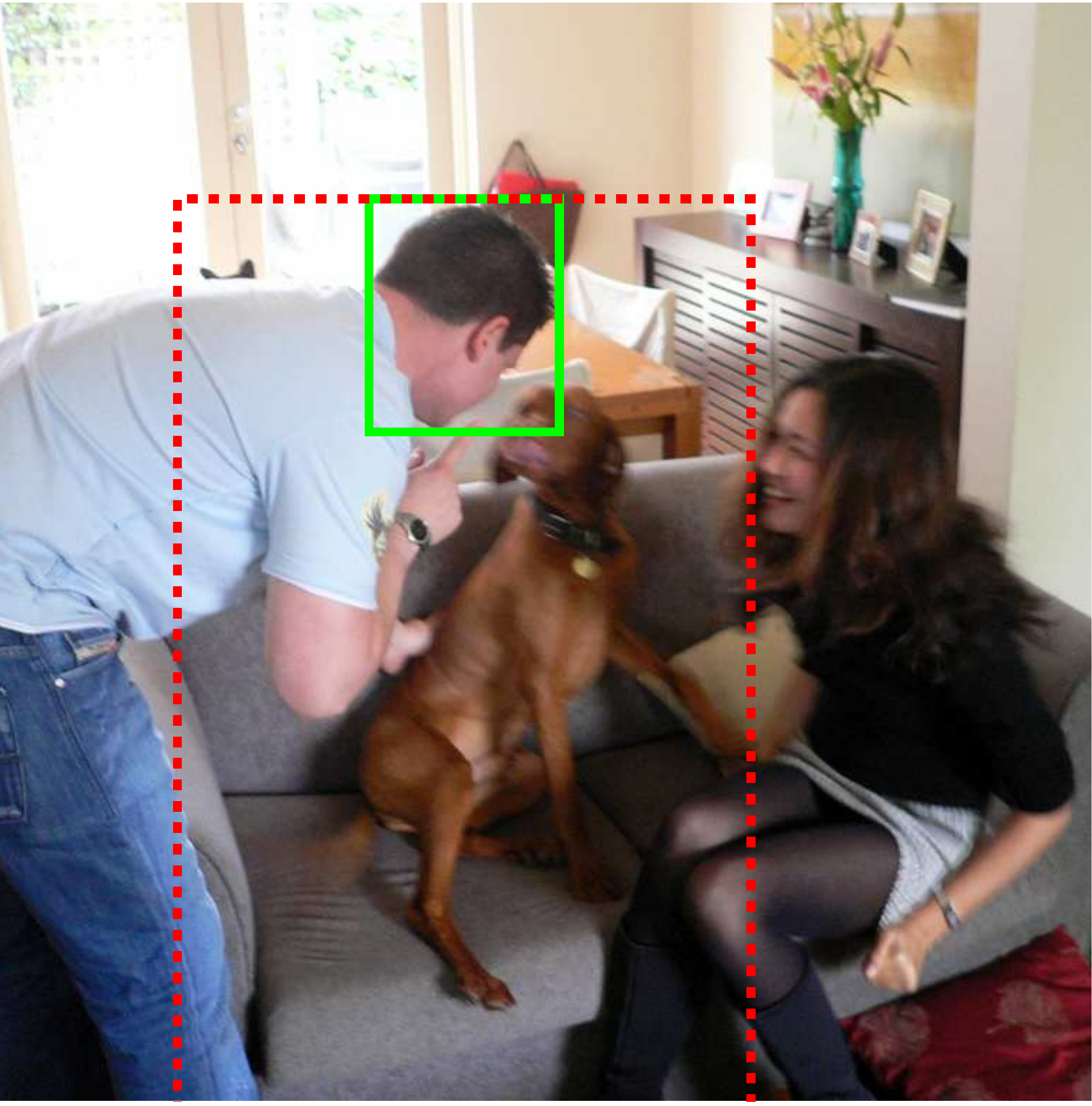}  
\includegraphics[height=0.21\linewidth]{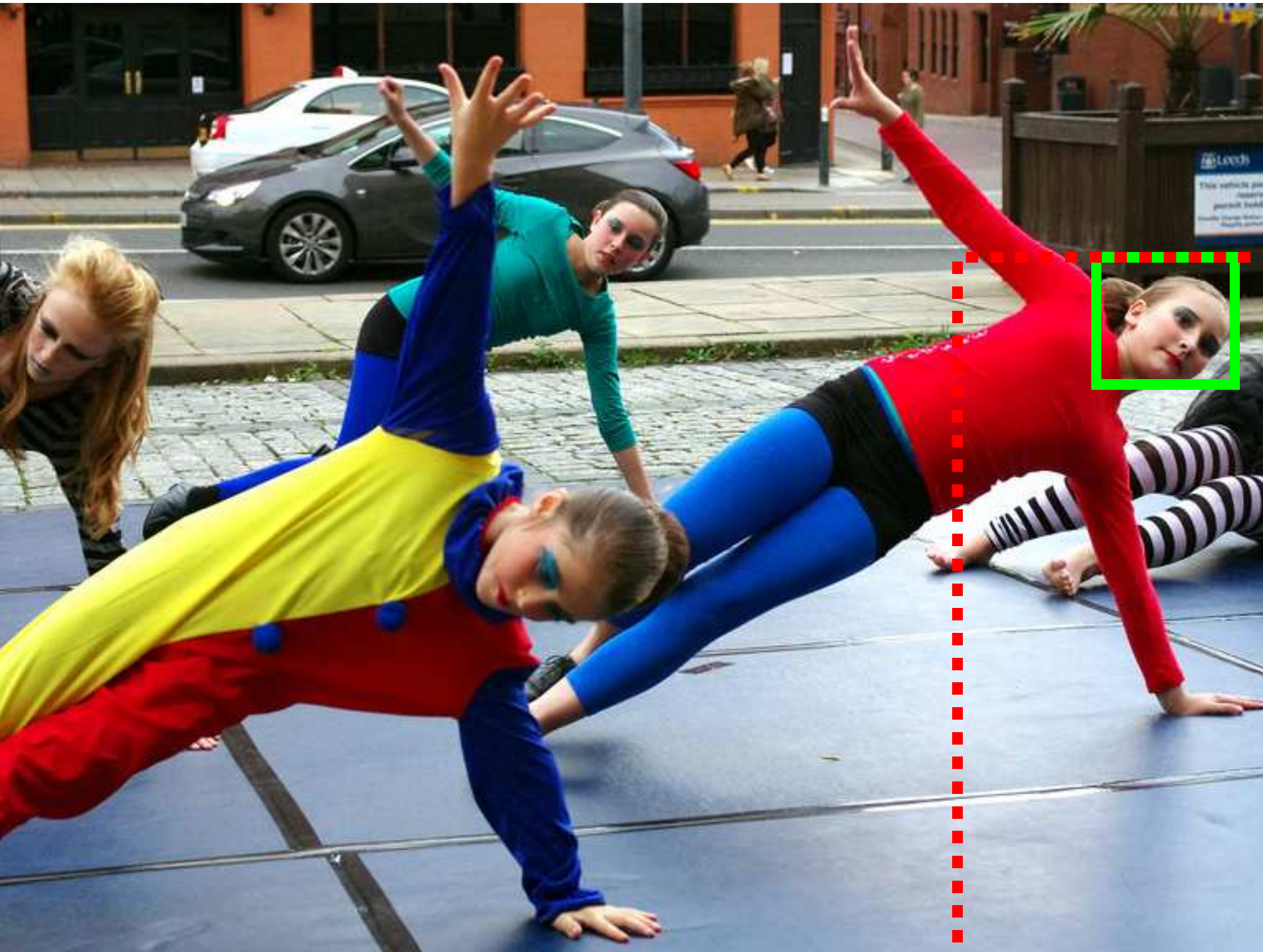}
\\
\includegraphics[height=0.21\linewidth]{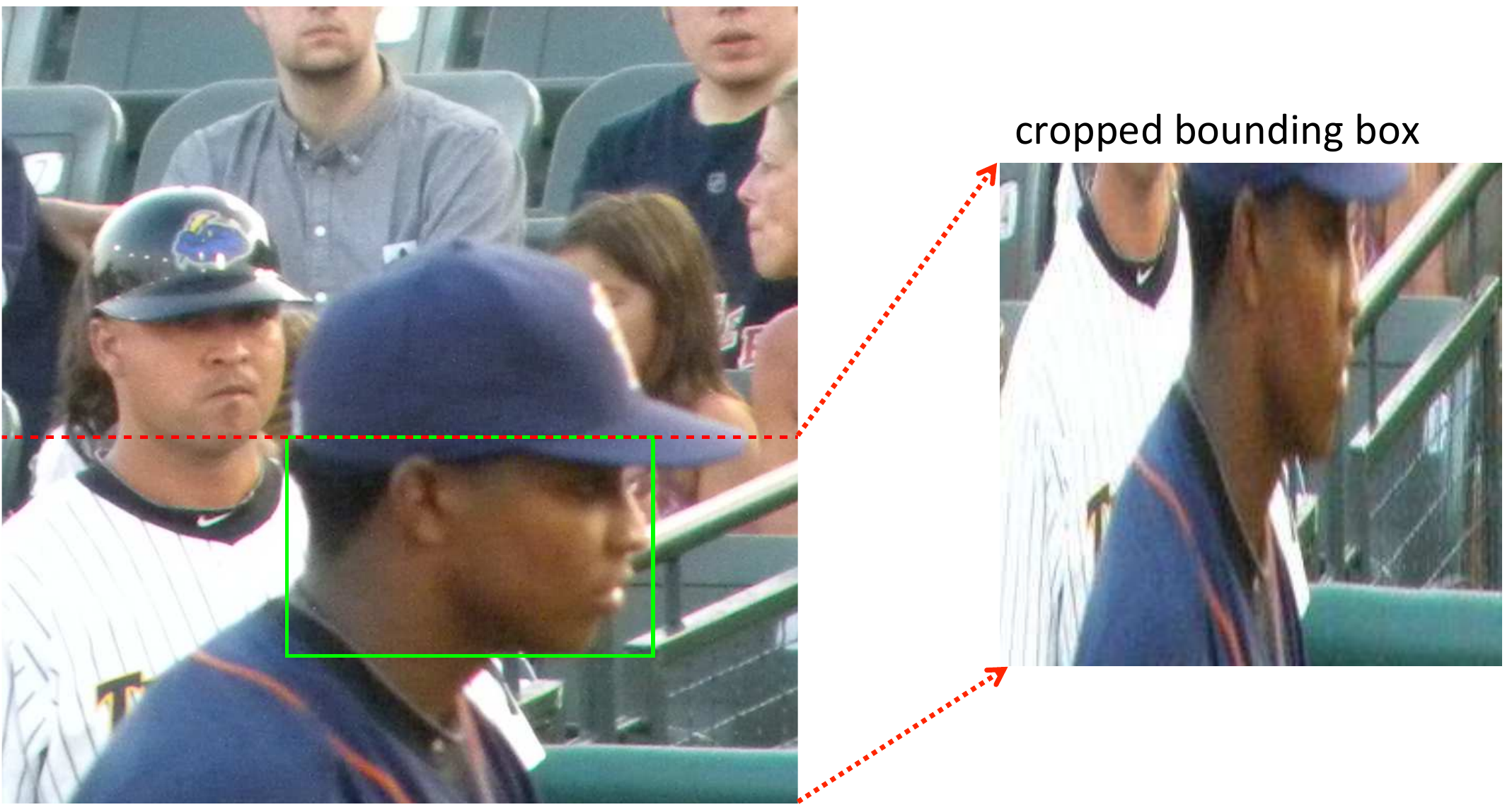}
&
\includegraphics[height=0.21\linewidth]{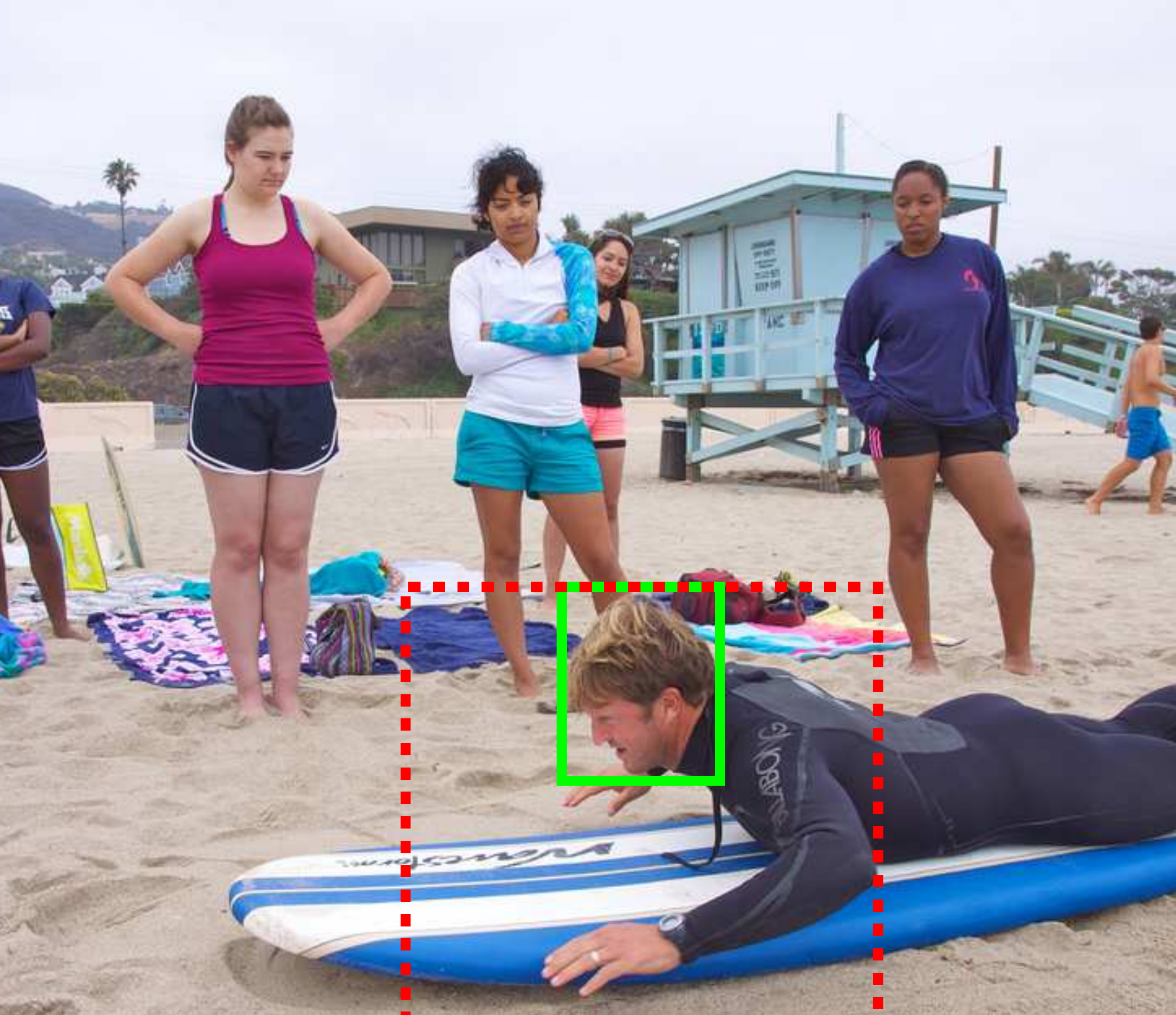}
\includegraphics[height=0.21\linewidth]{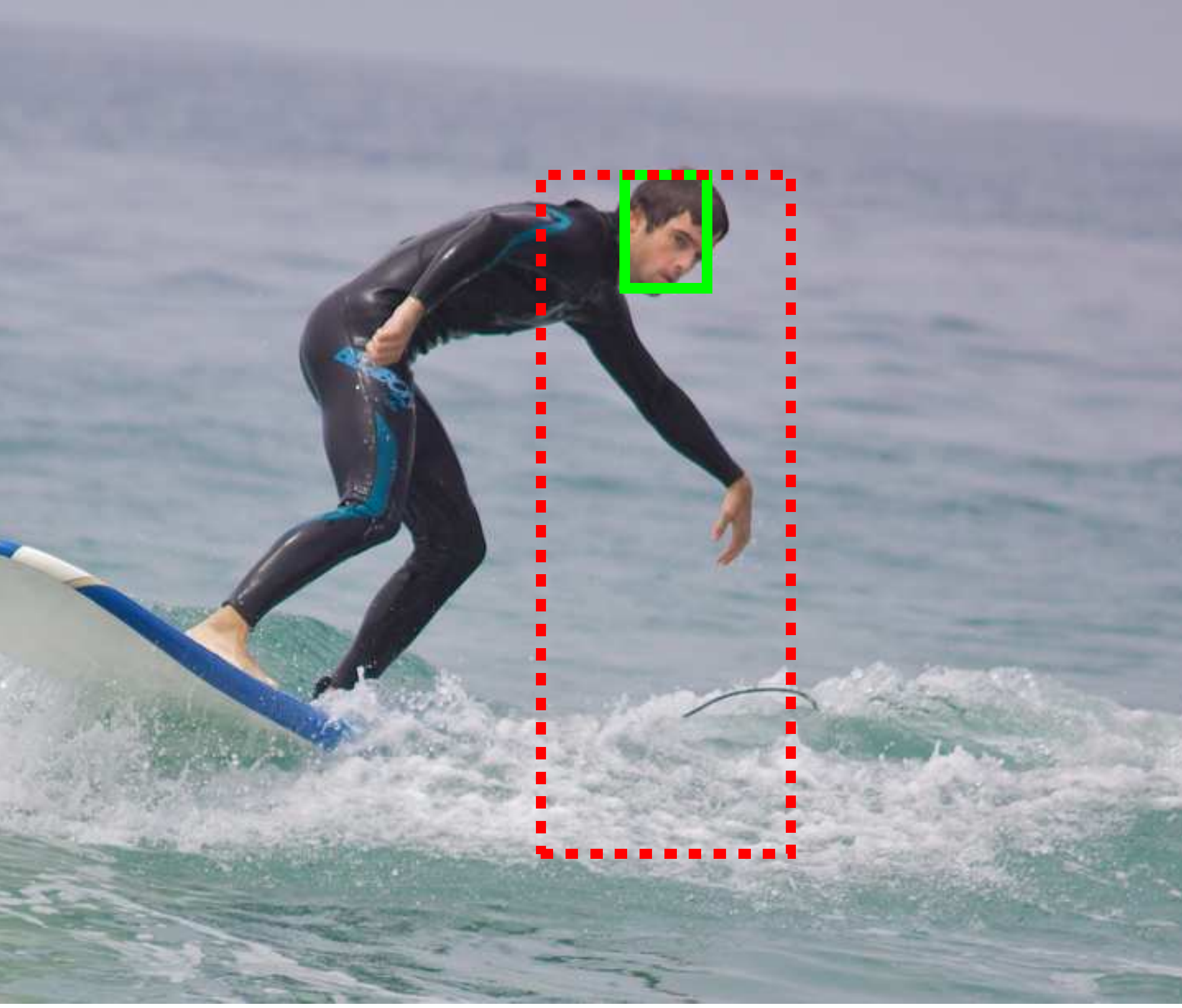}
\\
 (a) Body falls outside image & (b) Unusual pose \\
\includegraphics[height=0.21\linewidth]{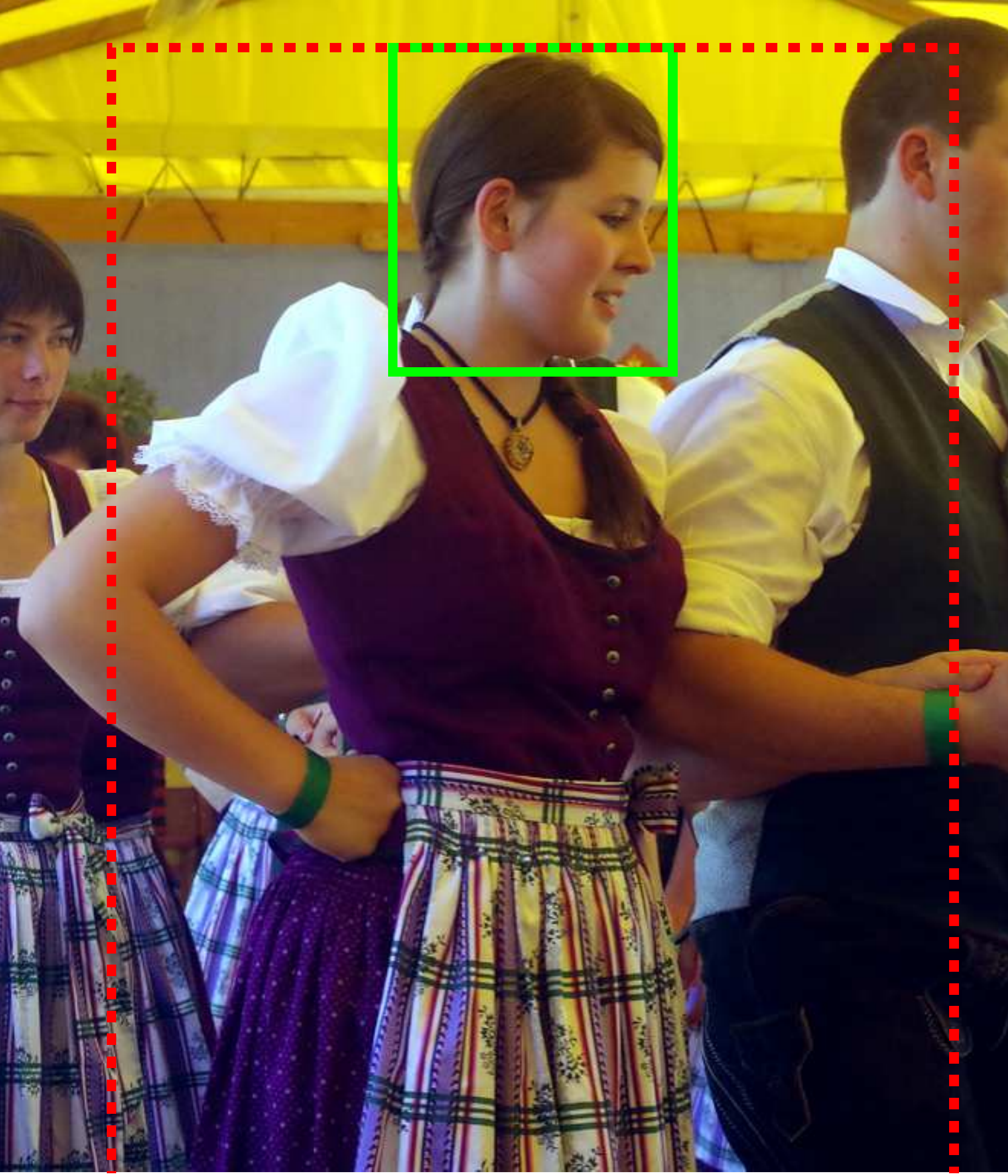} 
\includegraphics[height=0.21\linewidth]{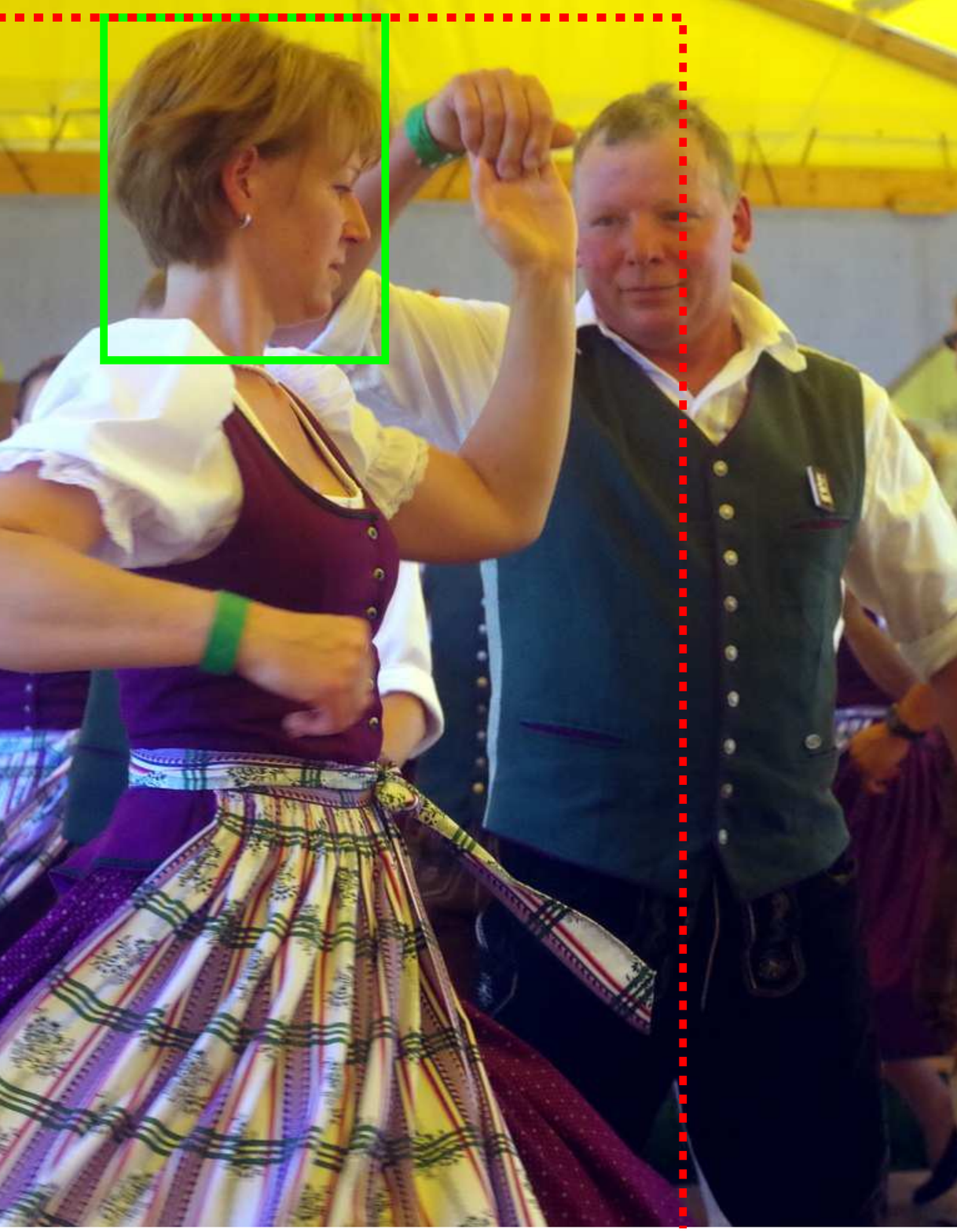}
&
\includegraphics[height=0.21\linewidth]{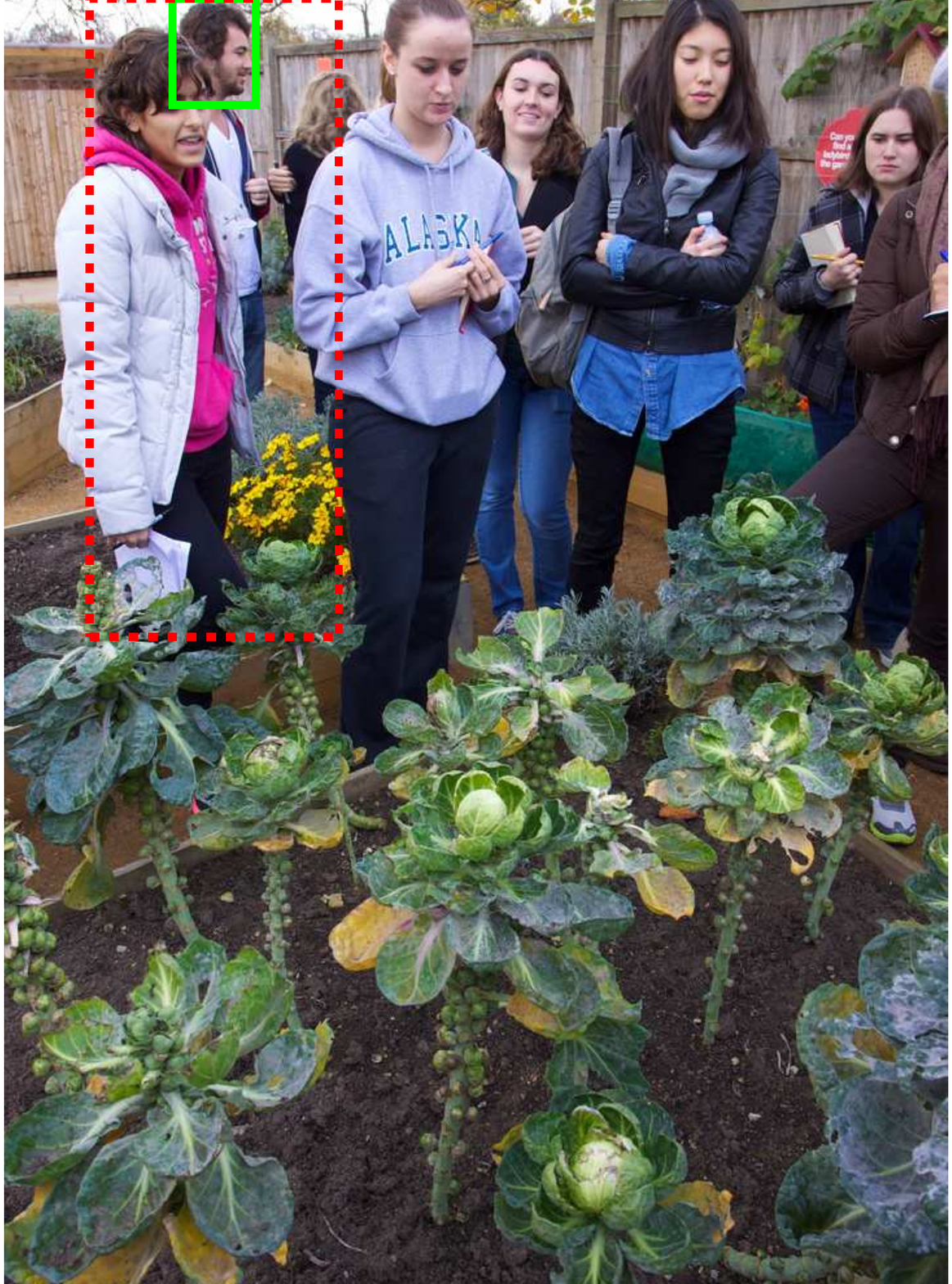}
\includegraphics[height=0.21\linewidth]{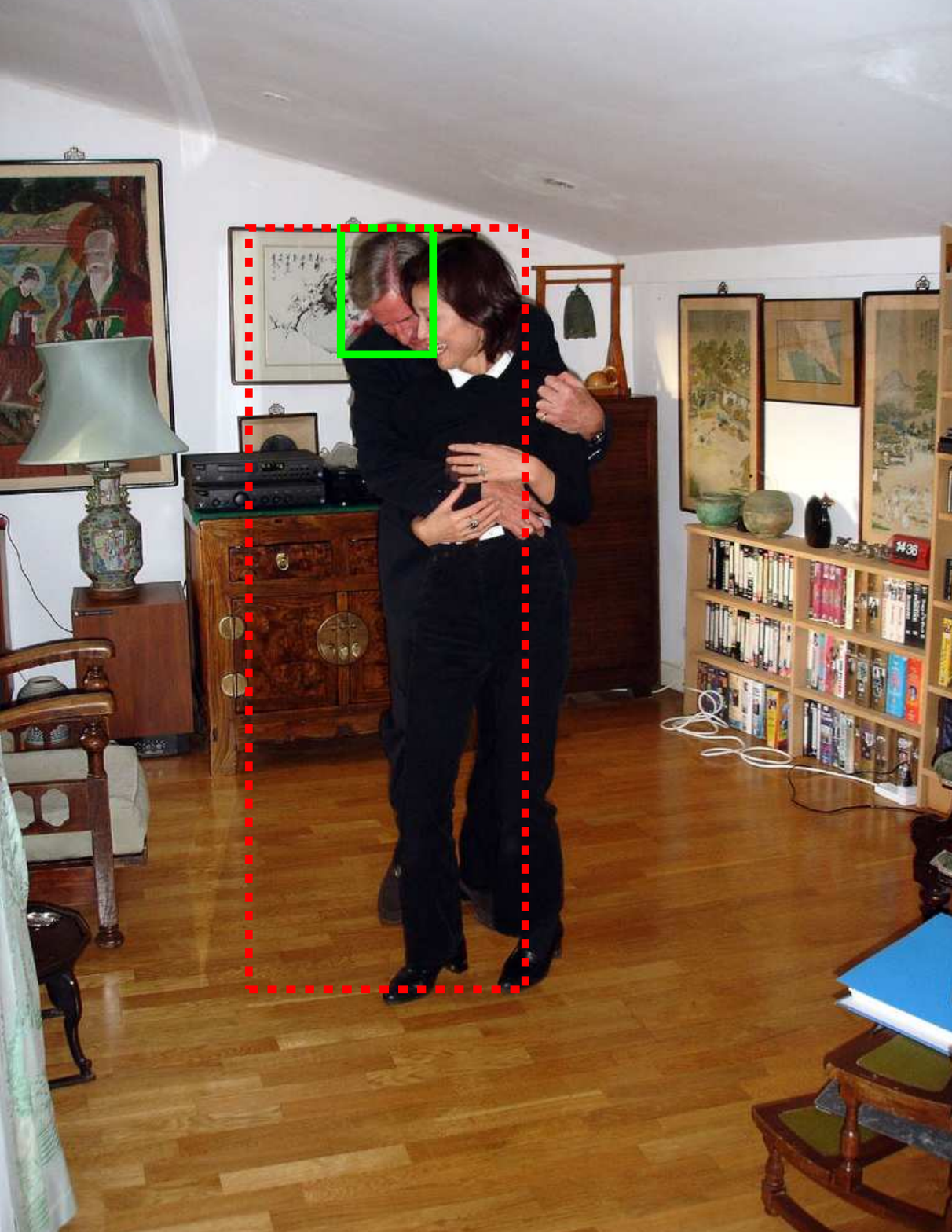}
\includegraphics[height=0.21\linewidth]{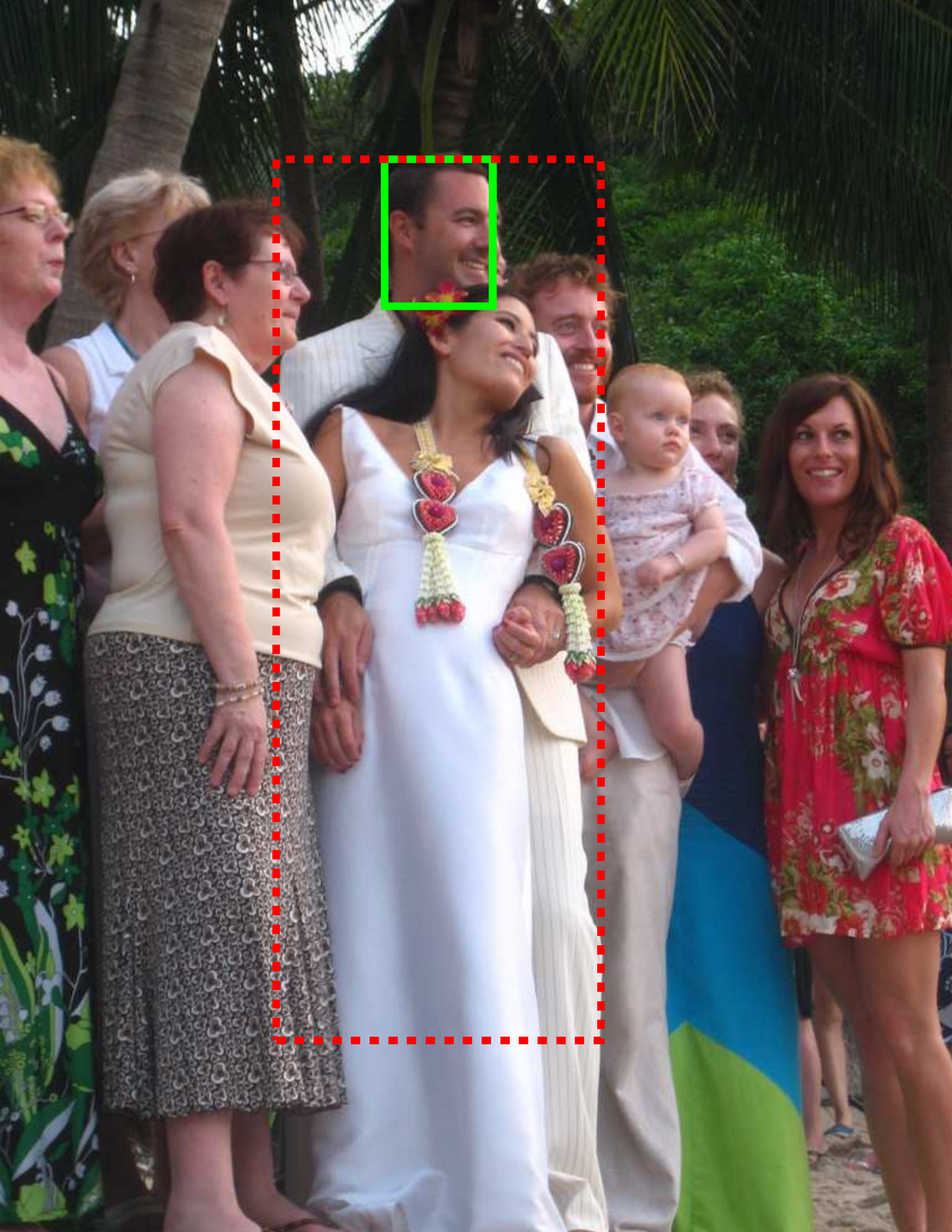} 
 \\
 (c) Same clothing similar pose &(d) Confusion with other person \\
\end{tabular}
\caption{Examples that the combination of the Global model and DeepFace misclassify and are recovered by using all of PIPER. \textbf{(a)} In a closeup shot the full body falls outside the image and the extracted full-body patch, shown on the right, is severely misaligned. A profile-face poselet should handle this case without misalignment. \textbf{(b)} In unusual pose the full body patch may fall on the background or \textbf{(d)} on another person which will further confuse the classifier. In \textbf{(c)} people have the same clothes and similar pose which will confuse the global model.}
\label{fig:correct_examples}
\end{figure*}

We first present experimental results on the person recognition task with our {\em PIPA} dataset. 
It is a standard supervised classification task as we train and test on same set of identities. Since the set of identities between training and test sets is disjoint, we split our test set in two equal subsets. We train an SVM on the first, use it to compute $P_i(y|X)$ on the second and vice versa. We then use the weights $w$ trained on the validation set to get our final prediction as the identity that maximizes the score in equation~\ref{eq:joint} and we average the accuracy from both halves of the test set. Qualitative examples are shown on Figure~\ref{fig:correct_examples}.

\subsubsection{Overall Performance}
Table \ref{table:classification} shows the accuracy in this setting compared to some standard baselines. We compared it against DeepFace~\cite{deepface}, which is one of the state-of-the-art face recognizers. Although it is very accurate, it is a frontal face recognizer, so it doesn't trigger on 48\% of our test set and we use chance performance for those instances. As a second baseline we trained an SVM using the FC7 features of the CNN proposed by Krizhevsky \etal. and pretrained on ImageNet\cite{Deng09imagenet:a}. We use its activations after showing it the full body image patch for each instance. The Global Model baseline is the same CNN, except it was fine-tuned for the task of identity recognition on our training set. We also tested the performance of our model by combining the sparse part predictions, i.e. using $\hat{P}_i(y|X)$ instead of $P_i(y|X)$ in equation~\ref{eq:joint}. The performance gap of more than 7\% shows that sparsity filling is essential to achieve high recognition accuracy.

\begin{table}[t]
\centering
\begin{tabular}{|c|c|}
\hline
Method&Classification accuracy\\
\hline
Chance Performance & 0.17\%\\
DeepFace \cite{deepface} & 46.66\%\\
FC7 of Krizhevsky \etal \cite{Krizhevsky12} &  56.07\%\\
Global Model &  67.60\% \\
PIPER w/out sparsity filling & 75.35\% \\
PIPER & 83.05\% \\
\hline
\end{tabular}
\caption{Person recognition results on PIPA test set using 6442 training examples over 581 identities} \label{table:classification}
\end{table}

\subsubsection{Ablation Study}

Our method consists of three components -- the fine-tuned Krizhevsky CNN (the Global Model), the DeepFace recognizer, and a collection of 108 Poselet-based recognizers. In this section we explore using all combinations of these three components\footnote{Since our method relies on sparsity filling from a global model $P_0(y|X)$, to remove the effect of the global model we simply set  $P_0(y|X)$ to be uniform distribution.}. For each combination we retrain the mixture weights $w$ and re-tune the hyper parameters. Table~\ref{table:ablation} shows the performance of each combination of these components. As the table shows, the three parts PIPER are complementary and combining them is necessary to achieve the best performance. 

\begin{table}[t]
\centering
\begin{tabular}{|c|c|c|c|}
\hline
Global Model &  DeepFace\cite{deepface} &  Poselets  & Accuracy\\
\hline
\checkmark & -- & -- & 67.60\%\\
--&\checkmark&--& 46.66\%\\
--&-- & \checkmark & 72.18\%\\
\checkmark & \checkmark &--& 79.95\%\\
\checkmark& --& \checkmark &  78.79\% \\
--& \checkmark & \checkmark & 78.08\% \\
\checkmark & \checkmark & \checkmark & 83.05\%\\
\hline
\end{tabular}
\caption{Person recognition performance on the PIPA test set using 6442 training examples over 581 identities as we disable some of the components of our method. PIPER gets more than 3\% gain over the very strong baseline of using the fine-tuned CNN combined with the DeepFace model. DeepFace' s score is low because it only fires on 52\% of the test images and we use chance performance for the rest.}
\label{table:ablation}
\end{table}

\subsubsection{Performance on face and non-face instances}

Since the presence of a high resolution frontal face provides a strong cue for identity recognition and allows us to use the face recognizer, it is important to consider the performance when a frontal face is present and when it is not. Table~\ref{table:faces-nofaces} shows the performance on the face and non-face part of our test set. We considered the instances for which DeepFace generated a signature as the face subset. As the figure shows, when the face is not present we can significantly outperform a fine-tuned CNN on the full image. More importantly, the contextual cues and combinations of many classifiers allow us to significantly boost the recognition performance even when a frontal face is present. 

\begin{table}[t]
\centering
\begin{tabular}{|c|c|c|}
\hline
Method & Non-faces subset & Faces subset \\
\hline
Global Model & 64.3\% & 70.6\% \\
DeepFace\cite{deepface} & 0.17\% & 89.3\% \\
PIPER & 71.8\% & 93.4\% \\
\hline
\end{tabular}
\caption{Performance on the test set when split into the subset where frontal face is visible (52\%) and when it is not (48\%).} \label{table:faces-nofaces}
\end{table}

\begin{figure}[t]
\centering
\includegraphics[width=0.8\linewidth]{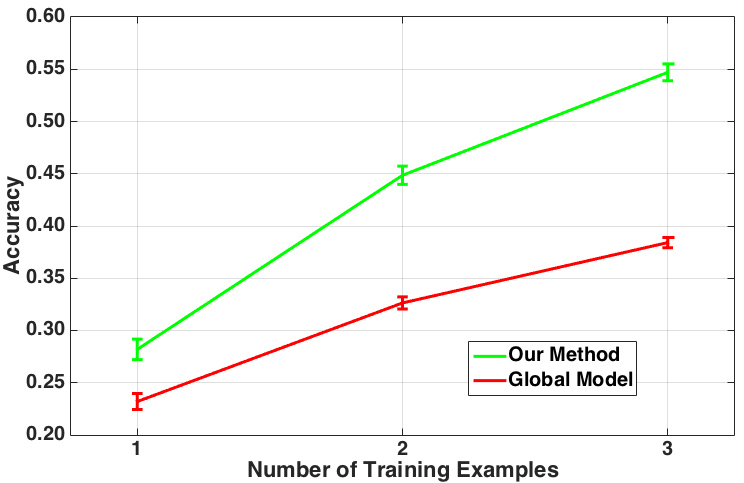}
\caption{Recognition accuracy as a function of number of training examples per identity, with $\sigma=1$ error bar. As we increase the number of training examples our system's accuracy grows faster than the full-body baseline. Chance performance is 0.0017.}
\label{fig:oneshot}
\end{figure}

\begin{figure}[t]
\centering
\includegraphics[width=0.8\linewidth]{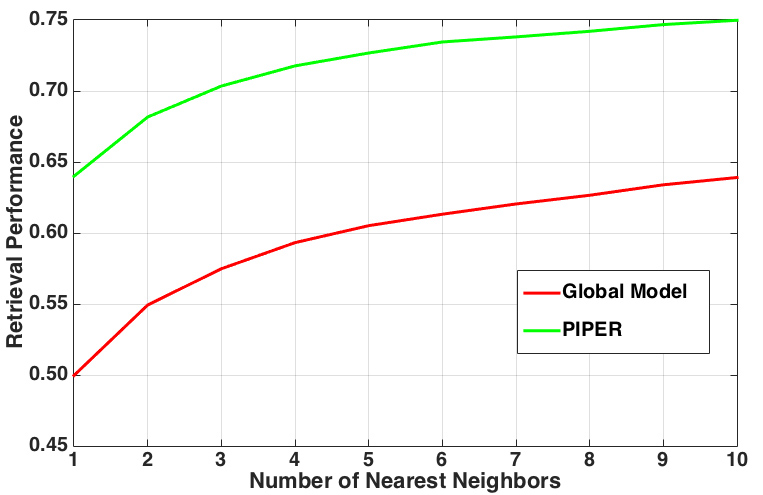}
\caption{Performance of our method on identity retrieval.}
\label{fig:unsupervised}
\end{figure}

\subsection{One-Shot Learning}
Figure \ref{fig:oneshot} shows the performance of our system when the training set is tiny. We randomly pick one, two or three instances of each identity in our test set, train on those and report results on the rest of the test set. Our system performs very well even with a single training example per identity, achieving 28.1\% accuracy for our test set of 581 identities. This result illustrates the powerful generalization capability of our method. The generalization capabilities of deep features are well studied, but we believe here we are also helped by the mixture of multiple part-based classifiers, since our system improves faster than the global method of fine-tuned Krizhevsky's CNN.

\subsection{Unsupervised identity retrieval}
We evaluate our method on the task of retrieval: Given an instance, we measure the likelihood that one of the K nearest neighbors will have the same identity.

\begin{figure}\centering
Query~~~~~ Top 1 ~~~~ Top 2 ~~~~ Top 3 ~~~~ Top 4 ~~~~ Top 5 ~~~~  \\
\includegraphics[height=0.2\linewidth, width=0.15\linewidth]{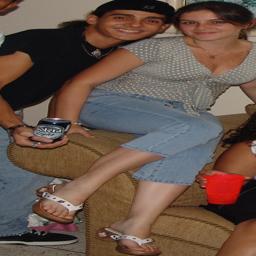} 
\includegraphics[height=0.2\linewidth, width=0.15\linewidth]{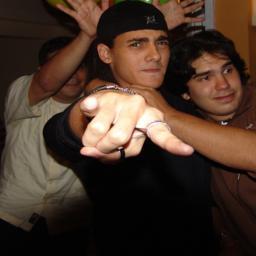} 
\includegraphics[height=0.2\linewidth, width=0.15\linewidth]{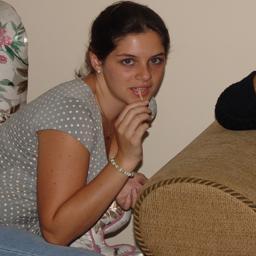}
\includegraphics[height=0.2\linewidth, width=0.15\linewidth]{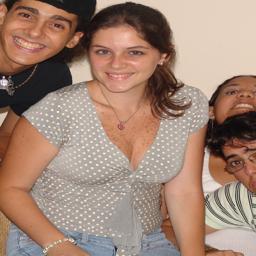} 
\includegraphics[height=0.2\linewidth, width=0.15\linewidth]{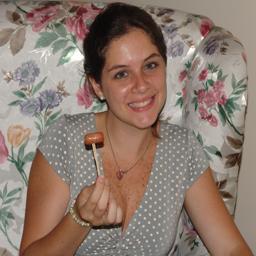} 
\includegraphics[height=0.2\linewidth, width=0.15\linewidth]{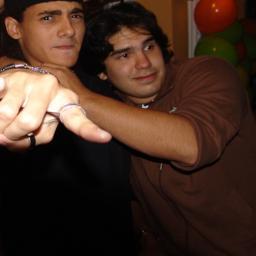} \\
\includegraphics[height=0.2\linewidth, width=0.15\linewidth]{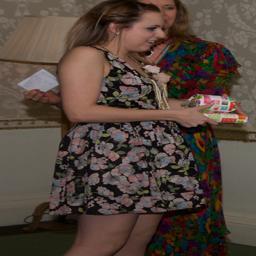} 
\includegraphics[height=0.2\linewidth, width=0.15\linewidth]{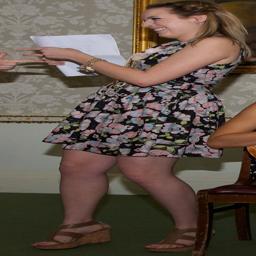} 
\includegraphics[height=0.2\linewidth, width=0.15\linewidth]{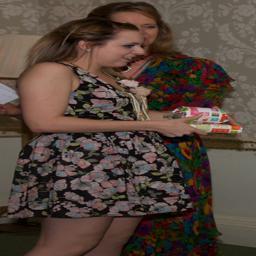} 
\includegraphics[height=0.2\linewidth, width=0.15\linewidth]{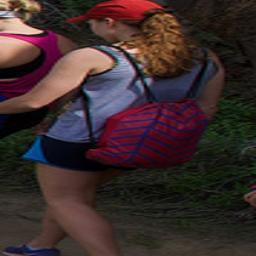} 
\includegraphics[height=0.2\linewidth, width=0.15\linewidth]{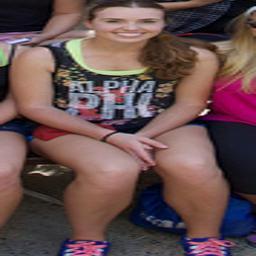} 
\includegraphics[height=0.2\linewidth, width=0.15\linewidth]{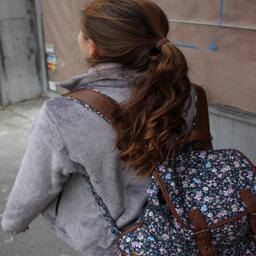} \\
\includegraphics[height=0.2\linewidth, width=0.15\linewidth]{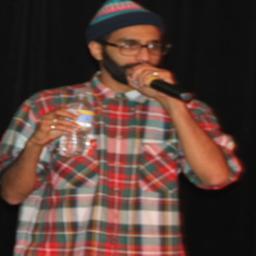} 
\includegraphics[height=0.2\linewidth, width=0.15\linewidth]{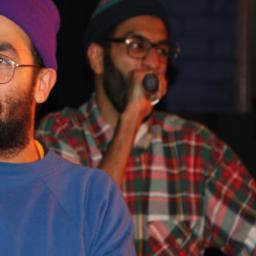} 
\includegraphics[height=0.2\linewidth, width=0.15\linewidth]{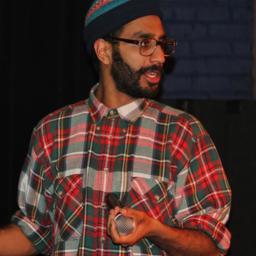} 
\includegraphics[height=0.2\linewidth, width=0.15\linewidth]{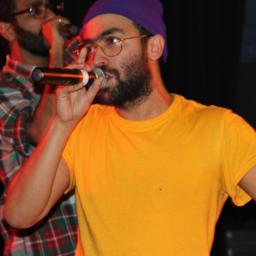}
\includegraphics[height=0.2\linewidth, width=0.15\linewidth]{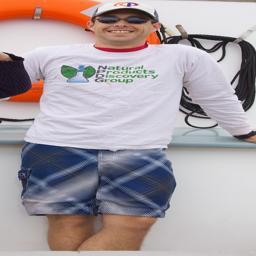}
\includegraphics[height=0.2\linewidth, width=0.15\linewidth]{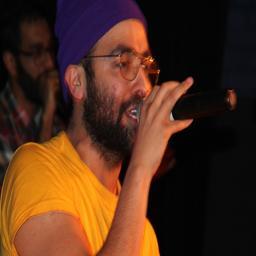} \\
\includegraphics[height=0.2\linewidth, width=0.15\linewidth]{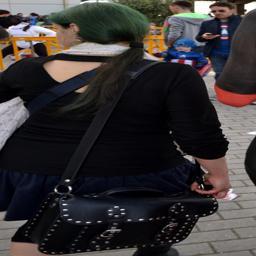} 
\includegraphics[height=0.2\linewidth, width=0.15\linewidth]{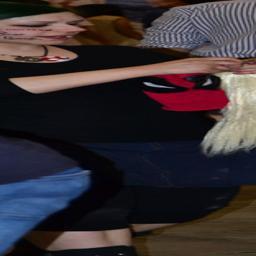} 
\includegraphics[height=0.2\linewidth, width=0.15\linewidth]{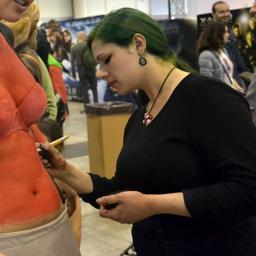} 
\includegraphics[height=0.2\linewidth, width=0.15\linewidth]{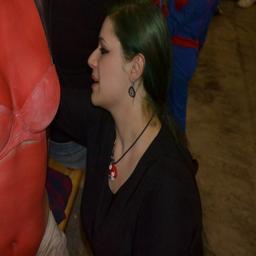} 
\includegraphics[height=0.2\linewidth, width=0.15\linewidth]{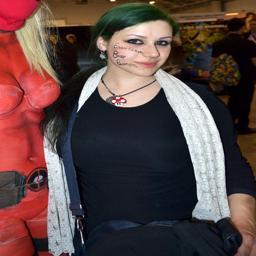} 
\includegraphics[height=0.2\linewidth, width=0.15\linewidth]{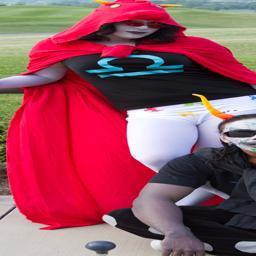} \\
\includegraphics[height=0.2\linewidth, width=0.15\linewidth]{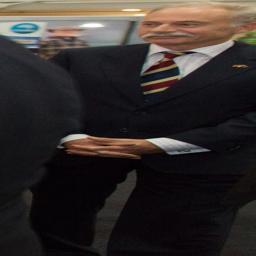} 
\includegraphics[height=0.2\linewidth, width=0.15\linewidth]{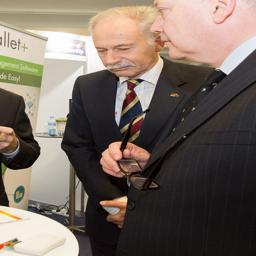} 
\includegraphics[height=0.2\linewidth, width=0.15\linewidth]{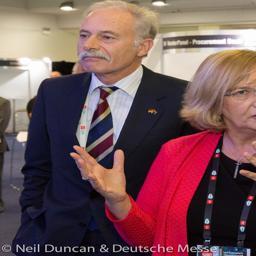} 
\includegraphics[height=0.2\linewidth, width=0.15\linewidth]{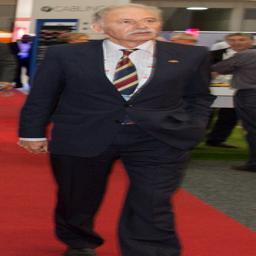} 
\includegraphics[height=0.2\linewidth, width=0.15\linewidth]{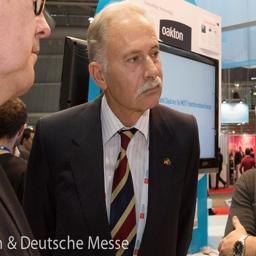} 
\includegraphics[height=0.2\linewidth, width=0.15\linewidth]{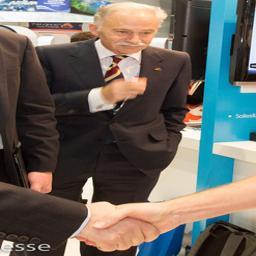} \\
\includegraphics[height=0.2\linewidth, width=0.15\linewidth]{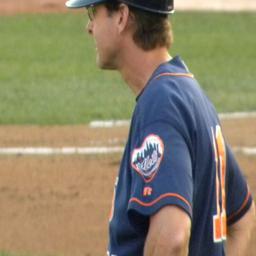} 
\includegraphics[height=0.2\linewidth, width=0.15\linewidth]{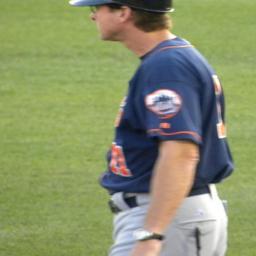} 
\includegraphics[height=0.2\linewidth, width=0.15\linewidth]{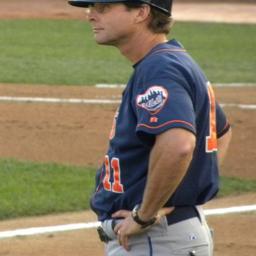} 
\includegraphics[height=0.2\linewidth, width=0.15\linewidth]{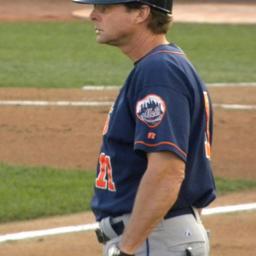} 
\includegraphics[height=0.2\linewidth, width=0.15\linewidth]{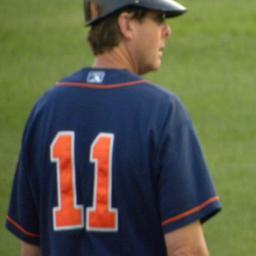} 
\includegraphics[height=0.2\linewidth, width=0.15\linewidth]{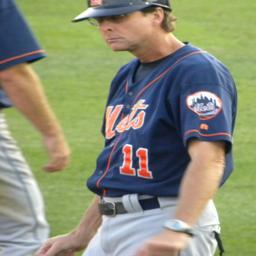} \\
\caption{Example of PIPER results on unsupervised identity retrieval. 
For each row, the left one is the query image, then the top 5 ranked retrieval images. Those are cropped bounding box images on test split. }
\label{fig:retrieval}
\end{figure}

To do this we used the SVMs trained on split 0 of the validation set to predict the 366 identities in the validation set. We applied them to the instances in the test set to obtain a 366-dimensional feature vector for each part and we combine the part predictions using equation~\ref{eq:joint} with $w$ trained on the validation set to obtain a single 366-dimensional feature for each instance in the test set.
We then, for each instance of the test set, compute the K nearest neighbors using Eucledian distance and we consider retrieval as successful if at least one of them is of the same identity. This has the effect of using the identities in the validation set as exemplars and projecting new instances based on their similarities to those identities. As Figure~\ref{fig:unsupervised} shows our method is quite effective on this task -- despite the low dimensional feature and without any metric learning, the nearest neighbor of 64\% of the examples is of the same identity as the query. If we use the predictions of the Krizhevsky's CNN trained on ImageNet and fine-tuned on our training set, which is known to be a very powerful baseline, the nearest neighbor is of the same class in only 50\% of the examples. This suggests that our model is capable of building a powerful and compact identity vector independent of pose and viewpoint. Examples of our method are shown in Figure \ref{fig:retrieval}.

\section{Conclusion}
We described PIPER, our method for viewpoint and pose independent person recognition. We showed that PIPER significantly outperforms our very strong baseline -- combining a state-of-the-art CNN system fine-tuned on our dataset with a state-of-the-art frontal face recognizer. PIPER can learn effectively even with a single training example and performs surprisingly well at the task of image retrieval. While we have used PIPER for person recognition, the algorithm readily applies to generic instance co-identification, such as finding instances of the same car or the same dog. We introduced the {\em People In Photo Albums} dataset, the first of its kind large scale data set for person coidentification in photo albums. 
We hope our dataset will steer the vision community towards the very important and largely unsolved problem of person recognition in the wild. 

\paragraph{Acknowledgments} 
We would like to thank Nikhil Johri for collecting dataset and developing the annotation interfaces. We also would like to thank Trevor Darrell for useful discussions on the ideas of the paper. 

\paragraph{Copyrights}
The copyright references of all the images shown in the paper are in \url{http://www.eecs.berkeley.edu/~nzhang/piper_image_license.pdf}.
\balance
{\small
\bibliographystyle{ieee}
\bibliography{bibliography}
}

\end{document}